\documentclass[10pt,twocolumn,letterpaper]{article}

\usepackage[pagenumbers]{wacv} %

\usepackage{times}
\usepackage{epsfig}
\usepackage{graphicx}
\usepackage{amsmath}
\usepackage{amssymb}
\usepackage{graphicx}
\usepackage{amsmath}
\usepackage{amssymb}
\usepackage{booktabs}
\usepackage[toc,page]{appendix}
\usepackage{makecell}
\usepackage{siunitx}
\usepackage[version=4]{mhchem}
\usepackage{amssymb}%
\usepackage{pifont}%
\usepackage{makecell}
\usepackage{siunitx, mhchem}
\usepackage{booktabs}
\usepackage{multirow}
\usepackage{tikz}
\usepackage{tikzlings}
\usepackage{tikzducks}
\usepackage{fontawesome5}
\usepackage{svg}
\usepackage{environ}
\usepackage{etoolbox}
\usepackage{xspace}
\usepackage{xcolor}
\usepackage{booktabs,colortbl,tabularx}
\newtoggle{arxiv}

\NewEnviron{arxivPrint}{%
  \iftoggle{arxiv}{\BODY}{\xspace}%
}
\newtoggle{conf} 
\NewEnviron{confPrint}{%
  \iftoggle{conf}{\BODY}{\xspace}%
}

\usepackage[export]{adjustbox}
\usepackage{colortbl}
\usepackage{subcaption}
\usepackage{enumitem}
\usepackage{booktabs}

\usepackage[pagebackref=true,breaklinks=true,letterpaper=true,colorlinks,bookmarks=false]{hyperref}
\usepackage[capitalize]{cleveref}
\crefname{section}{Sec.}{Secs.}
\crefname{table}{Tab.}{Tabs.}
\crefname{equation}{Eq.}{Eqs.}
\crefname{figure}{Fig.}{Figs.}
\Crefname{section}{Sec.}{Secs.}
\Crefname{table}{Tab.}{Tabs.}
\Crefname{equation}{Eq.}{Eqs.}
\Crefname{figure}{Fig.}{Figs.}

\def\wacvPaperID{338} %
\def\confName{WACV}
\def\confYear{2024}

\begin{document}

\title{FakeOut: Leveraging Out-of-domain Self-supervision \\for Multi-modal Video Deepfake Detection}

\author{Gil Knafo\\
Reichman University\\
{\tt\small knafo.gil@post.runi.ac.il}
\and
Ohad Fried\\
Reichman University\\
{\tt\small ofried@runi.ac.il}
}
\maketitle
\toggletrue{arxiv}
\begin{abstract}
\label{abstract}
   Video synthesis methods rapidly improved in recent years, allowing easy creation of synthetic humans. This poses a problem, especially in the era of social media, as synthetic videos of speaking humans can be used to spread misinformation in a convincing manner.
   Thus, there is a pressing need for accurate and robust deepfake detection methods,
   that can detect forgery techniques not seen during training.
   In this work, we explore whether this can be done by leveraging a multi-modal, out-of-domain backbone trained in a self-supervised manner, adapted to the video deepfake domain. We propose FakeOut; a novel approach that relies on multi-modal data throughout both the pre-training phase and the adaption phase.
   We demonstrate the efficacy and robustness of FakeOut in detecting various types of deepfakes, especially manipulations which were not seen during training.
   Our method achieves state-of-the-art results in cross-dataset generalization on audio-visual datasets. This study shows that, perhaps surprisingly, training on out-of-domain videos (i.e., not especially featuring speaking humans), 
   can lead to better deepfake detection systems. \begin{arxivPrint} Code is available on GitHub\footnote{\url{https://github.com/gilikn/FakeOut}}.\end{arxivPrint}
\end{abstract}
\vspace*{-5mm}
\newcommand{\betweencellpdf}{\ensuremath{h^c}}
\newcommand{\betweencellpdffine}{\ensuremath{h^f}}
\newcommand{\approxbetweencellpdffine}{\ensuremath{\hat{h}^f}}
\newcommand{\incellpdf}{\ensuremath{f}}
\newcommand{\incellpdfnormalized}{\ensuremath{f'}}

\newcommand{\incellcdf}{\ensuremath{F}}

\newcommand{\totalpdf}{\ensuremath{f_{dd}}}
\newcommand{\totalcdf}{\ensuremath{F_{dd}}}

\newcommand{\ignorethis}[1]{}
\newcommand{\redund}[1]{#1}

\newcommand{\apriori    }     {\textit{a~priori}}
\newcommand{\aposteriori}     {\textit{a~posteriori}}
\newcommand{\perse      }     {\textit{per~se}}
\newcommand{\naive      }     {{na\"{\i}ve}}
\newcommand{\Naive      }     {{Na\"{\i}ve}}
\newcommand{\Identity   }     {\mat{I}}
\newcommand{\Zero       }     {\mathbf{0}}
\newcommand{\Reals      }     {{\textrm{I\kern-0.18em R}}}
\newcommand{\isdefined  }     {\mbox{\hspace{0.5ex}:=\hspace{0.5ex}}}
\newcommand{\texthalf   }     {\ensuremath{\textstyle\frac{1}{2}}}
\newcommand{\half       }     {\ensuremath{\frac{1}{2}}}
\newcommand{\third      }     {\ensuremath{\frac{1}{3}}}
\newcommand{\fourth     }     {\ensuremath{\frac{1}{4}}}

\newcommand{\Lone} {\ensuremath{L_1}}
\newcommand{\Ltwo} {\ensuremath{L_2}}

\newcommand{\mat        } [1] {{\text{\boldmath $\mathbit{#1}$}}}
\newcommand{\Approx     } [1] {\widetilde{#1}}
\newcommand{\change     } [1] {\mbox{{\footnotesize $\Delta$} \kern-3pt}#1}

\newcommand{\Order      } [1] {O(#1)}
\newcommand{\set        } [1] {{\lbrace #1 \rbrace}}
\newcommand{\floor      } [1] {{\lfloor #1 \rfloor}}
\newcommand{\ceil       } [1] {{\lceil  #1 \rceil }}
\newcommand{\inverse    } [1] {{#1}^{-1}}
\newcommand{\transpose  } [1] {{#1}^\mathrm{T}}
\newcommand{\invtransp  } [1] {{#1}^{-\mathrm{T}}}
\newcommand{\relu       } [1] {{\lbrack #1 \rbrack_+}}

\newcommand{\abs        } [1] {{| #1 |}}
\newcommand{\Abs        } [1] {{\left| #1 \right|}}
\newcommand{\norm       } [1] {{\| #1 \|}}
\newcommand{\Norm       } [1] {{\left\| #1 \right\|}}
\newcommand{\pnorm      } [2] {\norm{#1}_{#2}}
\newcommand{\Pnorm      } [2] {\Norm{#1}_{#2}}
\newcommand{\inner      } [2] {{\langle {#1} \, | \, {#2} \rangle}}
\newcommand{\Inner      } [2] {{\left\langle \begin{array}{@{}c|c@{}}
                               \displaystyle {#1} & \displaystyle {#2}
                               \end{array} \right\rangle}}

\newcommand{\twopartdef}[4]
{
  \left\{
  \begin{array}{ll}
    #1 & \mbox{if } #2 \\
    #3 & \mbox{if } #4
  \end{array}
  \right.
}

\newcommand{\fourpartdef}[8]
{
  \left\{
  \begin{array}{ll}
    #1 & \mbox{if } #2 \\
    #3 & \mbox{if } #4 \\
    #5 & \mbox{if } #6 \\
    #7 & \mbox{if } #8
  \end{array}
  \right.
}

\newcommand{\len}[1]{\text{len}(#1)}

\newlength{\w}
\newlength{\h}
\newlength{\x}

\definecolor{darkred}{rgb}{0.7,0.1,0.1}
\definecolor{darkgreen}{rgb}{0.1,0.6,0.1}
\definecolor{cyan}{rgb}{0.7,0.0,0.7}
\definecolor{otherblue}{rgb}{0.1,0.4,0.8}
\definecolor{maroon}{rgb}{0.76,.13,.28}
\definecolor{burntorange}{rgb}{0.81,.33,0}

\ifdefined\ShowNotes
  \newcommand{\colornote}[3]{{\color{#1}\textbf{#2} #3\normalfont}}
\else
  \newcommand{\colornote}[3]{}
\fi

\newcommand {\todo}[1]{\colornote{cyan}{TODO}{#1}}
\newcommand {\ohad}[1]{\colornote{burntorange}{OF:}{#1}}
\newcommand {\gili}[1]{\colornote{darkgreen}{GK:}{#1}}

\newcommand {\reqs}[1]{\colornote{red}{\tiny #1}}

\newcommand {\new}[1]{\colornote{red}{#1}}

\newcommand*\rot[1]{\rotatebox{90}{#1}}

\newcommand {\newstuff}[1]{#1}

\newcommand\todosilent[1]{}

\newcommand{\woBGmask}{{w/o~bg~\&~mask}}
\newcommand{\woMask}{{w/o~mask}}

\providecommand{\keywords}[1]
{
  \textbf{\textit{Keywords---}} #1
}

\newcommand {\shortcite}[1]{\cite{#1}}

\section{Introduction}
\label{sec:intro}

Fake renderings of humans have become easier to produce given recent advances in deep learning and off-the-shelf tools. Also known as deepfakes, these videos and images can be used to spread misinformation and to create convincing duplicates of real humans without their permission. Deepfake generation methods, driven mostly by deep generative models, are progressing rapidly.  To combat this phenomenon, many studies have been conducted and developed various types of deepfake detection techniques \cite{dang2020detection, fung2021deepfakeucl, gu2021spatiotemporal, haliassos2021lips, haliassos2022leveraging} and have been tested on many datasets and benchmarks \cite{chen2022ai, rossler2018faceforensics, li2020celeb, dolhansky2020deepfake, korshunov2018deepfakes, khalid2021fakeavceleb, korshunov2018deepfakes}. Although many detectors can perform well on in-distribution forgeries and manipulations, i.e., seen during training, performance plummets 
significantly on deepfakes generated by out-of-distribution manipulation techniques \cite{chai2020makes, haliassos2022leveraging, zhu2021face, haliassos2021lips}.

Most deepfake video detection methods are trained in a supervised manner \cite{agarwal2020detecting, haliassos2021lips, li2020face, zhou2021joint}, requiring immense amounts of annotated data, e.g. lip reading datasets such as Lip Reading in the Wild (LRW) \cite{Chung16}, besides the deepfake detection annotated datasets. The reliance on annotated data hinders their ability to make use of the vast amount of unlabelled videos available online. Other approaches, that use self-supervised techniques to leverage unlabelled data \cite{haliassos2022leveraging, cheng2022voice}, rely on facial and verbal data for pre-training. Although this data is indeed unlabelled, hence more available, it still requires a considerable effort of pre-processing during the self-supervised phase, such as face detection and lips region extraction. Driven by these drawbacks, in this work we focus on using a backbone that does not require any complex visual pre-processing for its self-supervised pre-training phase.
\begin{figure}
  \centering
   \includegraphics[width=\linewidth]{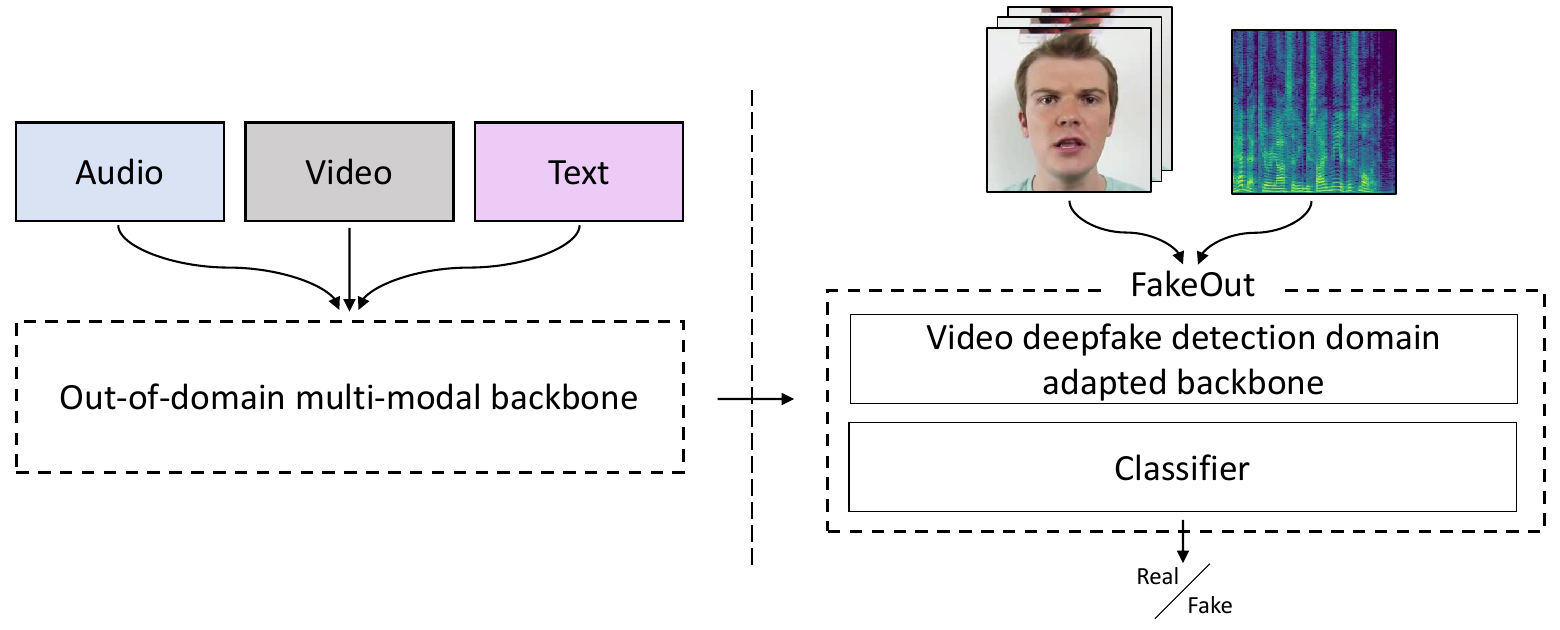}
   \vspace*{-4mm}
   \caption{\textbf{FakeOut schematic overview.} Adaption of out-of-domain self-supervised backbone to the video deepfake domain. FakeOut achieves state-of-the-art results without the reliance on extra facial datasets besides in-domain deepfake datasets for the fine-tuning phase.}
   \label{fig:preview-figure}
   \vspace*{-5mm}
\end{figure}
Furthermore, most of the methods that tackle the deepfake detection task rely only on one modality, usually the visual modality, and neglect the use of the auditory modality \cite{haliassos2021lips, rossler2019faceforensics++, wang2020cnn}. Studies from various domains show a significant benefit when incorporating multi-modal data and approaches \cite{alayrac2020self, shi2022avhubert}, but we believe it is still under-explored in the field of video deepfake detection.

Given the above, we propose our multi-modal approach that consists of a self-supervised phase and a supervised fine-tuning phase. The backbone in the first phase was pre-trained in a self-supervised manner on out of the deepfake domain examples. The second phase uses labelled data from video deepfake detection datasets for the fine-tuning process. Using a self-supervised backbone unrelated to human facial features emphasizes the  potential of studying deepfake detection systems with task-invariant backbones. We hope our approach will help tackle the proliferation of deepfakes, a crucial task in the era of social-media. 

Therefore, we present three main contributions:
\begin{enumerate}
    \vspace{-2mm}
   \item \emph{FakeOut} --- a multi-modal, video deepfake detection approach, which consists of two phases: (1) a backbone leveraging out-of-domain multi-modal videos to create robust representations for each modality, and (2) a fine-tuned task-specific detector, which achieves state-of-the-art performance in generalization tasks on audio-visual deepfake datasets.
   \vspace{-2mm}
   \item A robust end-to-end \textit{video face-tracking pipeline} supporting multi-person scenarios suitable for the deepfake detection field.
   \vspace{-2mm}
   \item \textit{Enriched datasets} for deepfake detection training, with an additional auditory modality.
\end{enumerate}

\section{Related work}
\label{sec:related work}
Many studies have been conducted in the field of deepfake detection since the beginning of the social media era and due to the broadly available deepfake generation methods.
Recently, self-supervised methods achieved a significant role in representation learning, which is beneficial for visual recognition tasks. We describe the recent advances in self-supervised learning, specifically regarding multi-modal approaches, and then delve into the facial forgery field.

\subsection{Multi-modal self-supervised learning}
\label{subsec:self-supervised-learning}

Self-supervised methods have been proposed for various computer-vision tasks, some with a multi-modal approach \cite{alayrac2020self, shi2022avhubert}. One significant work in this field was published by Alayrac \etal. \cite{alayrac2020self}, in which representations are learned in a self-supervised manner using unlabelled multi-modal videos. Their approach created a backbone, that ingests one or more modalities and outputs a representation for each modality of the video to enable versatile downstream tasks. 
In another closely related field, Shi~\etal.~\cite{shi2022avhubert} presented an approach for audio-visual speech representation learning, driven by a masked prediction cluster assignment task.
Self-supervision also advanced the ability of models to generalize, both in the deepfake detection task \cite{haliassos2022leveraging} and in other domains \cite{carlucci2019domain, radford2021learning}. In our work, we use the insights and the backbones provided by Alayrac \etal to achieve the goal of deepfake detection.

\subsection{Face forgery detection}
\label{subsec:forgery-face-detection}

\noindent
\textbf{Visual deepfake detection.}
Many visual approaches to detect deepfakes have been proposed \cite{fung2021deepfakeucl, wang2020cnn, haliassos2021lips, siegel2021media, gu2021spatiotemporal, heo2021deepfake, bai2023aunet, zhao2021learning}. Some methods are fully supervised \cite{haliassos2021lips} and some rely on data augmentations \cite{wang2020cnn}. 
Visual deepfake detection methods that leverage unannotated data have focused on deepfake imagery, for instance, Fung \etal \cite{fung2021deepfakeucl} proposed an unsupervised contrastive learning process, coupled with a downstream task of imagery deepfake detection. This work shows that contrastive learning pre-training can be relevant for image deepfake detection, and we expand this insight to the video domain. Another recent work in the visual field, is by Gu \etal \cite{gu2021spatiotemporal} which focuses on the spatio-temporal features of fake and real videos. They showed that temporal inconsistency may serve as an efficient indicator for video deepfake detection. Bai \etal \cite{bai2023aunet} proposed an approach for learning relations between action units for forgery detection, based on a supervised backbone and attention maps following the work by Li \etal \cite{li2017eac}. Another method by Zhao \etal \cite{zhao2021learning} incorporated both a supervised backbone and a self-supervised approach to provide patch-level annotations required for their training. Hand-crafted visual feature approaches were also discussed for deepfake detection \cite{siegel2021media}, but have yet to be explored jointly with neural-network approaches.

\noindent
\textbf{Multi-modal deepfake detection.} Recent studies explored multi-modal deepfake detection approaches \cite{mittal2020emotions,zhou2021joint, haliassos2022leveraging, chugh2020not, pu2021deepfake, cheng2022voice, khalid2021evaluation, cai2022you}. Zhou \etal \cite{zhou2021joint} proposed a joint audio-visual deepfake detection task and show that the synchronization between visual and auditory modalities could benefit deepfake detection. However, this method does not leverage unlabeled data and relies solely on annotated datasets. Another work leveraging multi-modality learning for video deepfake detection is by Haliassos \etal \cite{haliassos2022leveraging}, which tackles this task using audio and video modalities during the self-supervised pre-training phase alone, leaving aside the audio modality in the fine-tuning and inference stages. They used a non-contrastive learning approach to learn frame-level representations from real video examples only. Their method produces state-of-the-art results in generalization tasks, which shows the potential of multi-modality and self-supervision in the video deepfake detection field.
We leverage the multi-modal approach and expand it to affect both the pre-training phase, the fine-tuning phase and the inference phase --- via dataset enrichment, as we describe in \Cref{subsec:experiments-setup}. Cheng \etal \cite{cheng2022voice} and  Chugh \etal \cite{chugh2020not} also provide self-supervised multi-modal approaches. Cheng \etal used videos that feature human faces \cite{chung2018voxceleb2} during the pre-training phase. Our method seek to disentangle the coupling to pre-training on human faces, while not compromising performance. Chugh \etal proposed a supervised method, while not evaluating performance in the generalization tasks we focus on in \Cref{subsec:generalization-evaluation}.
\begin{figure*}[t]
    \includegraphics[width=1\textwidth]{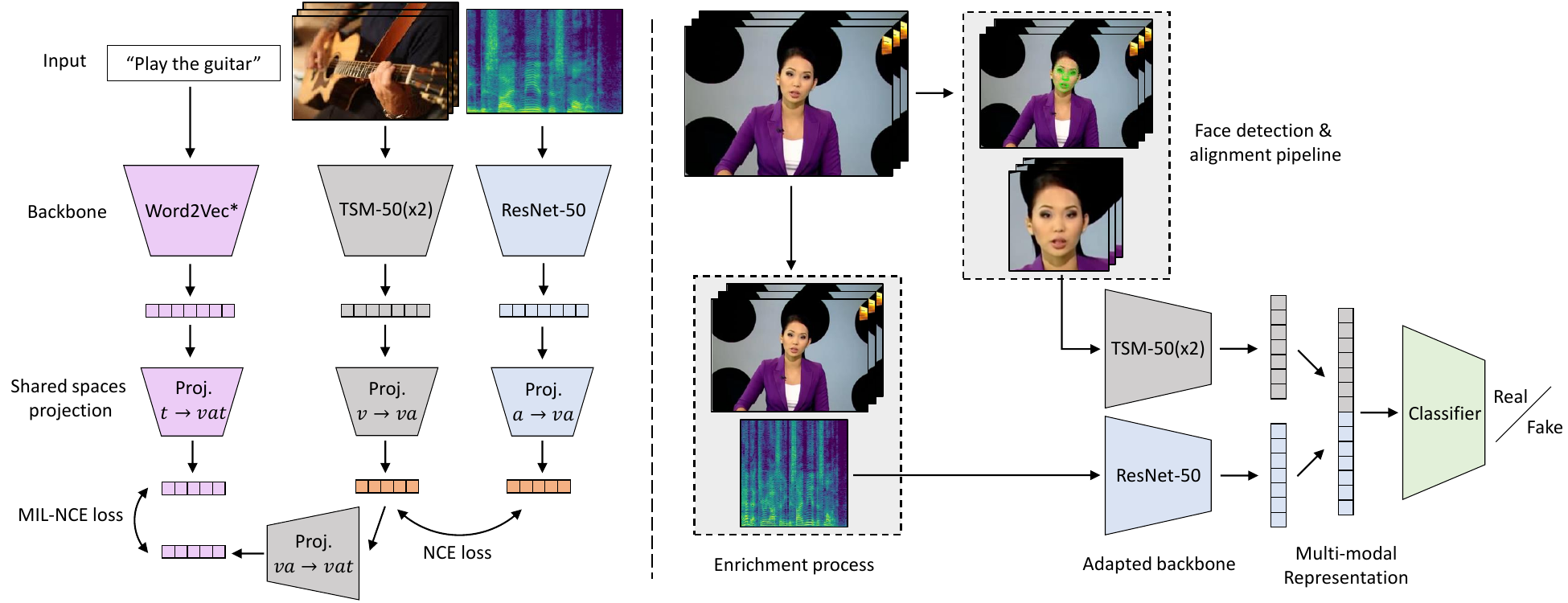}
    \centering
    \vspace*{-6mm}
    \caption{\textbf{Architecture of FakeOut --- Multi-Modal Learning System for Video Deepfake Detection.} Left: pre-training phase on multi-modal videos out of the deepfake domain, via MMV\cite{alayrac2020self}. Right: Adaptation phase, implemented using fine-tuning, in which we adapt the backbone to the video deepfake detection task. We utilize cross-modality representations of audio and video obtained by the face detection pipeline and the enrichment process.
    \label{fig:architecture}
    }
    \vspace*{-4mm}
\end{figure*}
\section{Method}
\label{sec:formatting}
Most video deepfake generation methods synthesize fake content using frame-by-frame manipulation techniques \cite{thies2019deferred, thies2016face2face, jiang2020deeperforensics}. Frame-by-frame techniques may fail to create temporally coherent results \cite{matern2019exploiting}. Therefore, visual representation learning can benefit from the sequential information embodied in the video. Particularly, it can learn temporal artifacts that can be exploited for deepfake detection \cite{gu2021spatiotemporal}. Another modality incorporated in videos is the audio modality, which can be utilized as well for detection when a mismatch between it and the visual modality occurs %
\cite{agarwal2020detecting}. Thus, we leverage these representations to create a joint representation for the detection task. 
To make the semantic representations more robust, we utilize a multi-modal backbone pre-trained on a large scale video dataset 
in a self-supervised manner. We take the pre-trained backbone that learned to output robust representations on out-of-facial-domain data and adapt it to the deepfake detection task.
\subsection{Backbone}
\label{subsec:backbone}
Our method is based on the \textit{``Self-Supervised MultiModal Versatile Networks"} (MMV) approach by Alayrac \etal \cite{alayrac2020self}, in which, modality specific representations are learned via contrastive-learning techniques. Below we reiterate the formulation of MMV, followed by a detailed explanation on how we adapt and extend their method for deepfake detection.

\noindent
\textbf{MMV representation learning setup.} Given a set of unlabelled videos with multiple modalities, MMV seeks to learn a model that can ingest any modality it was trained on and output a representation which can be compared to the other modalities.
Formally, in the implementation of Alayrac \etal, a video $x$ is an instance of a set of three modalities \(x = \{x_v, x_a, x_t\}\) where $x_v$, $x_a$, $x_t$ are the visual, audio and text modalities. These modalities concretely correspond to a sequence of RGB frames, audio samples, and discrete word tokens obtained by off-the-shelf Automatic Speech Recognition, respectively. A modality-specific backbone is a neural network $f_m$, (detailed in \Cref{subsec:experiments-setup}) that takes as input a modality of a video $x_m$, e.g., $x_v$, and outputs a representation vector of dimension \(d_m\). We denote the representation vector of modality $m$ by $z_m=f_m(x_m)$ and all modalities' representation vector by $z=f(x)$ where \(z = \{z_v, z_a, z_t\}\).
As MMV aims to enable easy comparison between representations of different modalities, that will also be used for loss calculations, representations are embedded, via projection heads, into a shared space $S_s \subset \mathbb{R}^{d_s}$, where $s$ contains the list of modalities that are embedded in the space, e.g. $s = va$ for a joint visual-audio space. Hence, modalities can be compared via a dot product. The vector representing the input modality $x_m$ in the space $S_s$ is denoted by $z_{m\rightarrow s}$. 
We use cross-modality representations to obtain the state-of-the-art results reported in \Cref{sec:experiments}.

\noindent
\textbf{MMV pre-training.} During the training of MMV, self-supervised tasks are constructed, which aim to align pairs of modalities. In order to construct positive training pairs across two modalities, two stream are sampled from the same location of a video. To construct negative training pairs, streams are sampled from two different videos. Contrastive loss is used to make these positive pairs similar and negative pairs dissimilar in their corresponding joint embedding space. Formally, given a video $x$, the following multi-modal contrastive loss was minimized:
\begin{equation}
  \mathcal{L}(x) = \lambda_{va}\mathrm{NCE}_{va}(x) + \lambda_{vt}\mathrm{MILNCE}_{vt}(x)  \label{eq:contrastiveloss}
  \vspace*{-1mm}
\end{equation}
where $\lambda_{mm'}$ indicates the weight of modality pair $m$ and $m'$. $\mathrm{NCE}_{va}(x)$ is the component corresponding to the visual-audio pair. It is defined by \Cref{eq:nce,eq:va}:
\begin{equation}
    \label{eq:nce}
    \mathrm{NCE}_{va}(x)\!=\!-log\left( \frac{\mathrm{VA}(f(x))}{\mathrm{VA}(f(x)) + \sum_{x'\sim\mathcal{N}(x)} \mathrm{VA}(f(x'))}\right)
\end{equation}
\begin{equation}
    \label{eq:va}
    \mathrm{VA}(z) = \exp(z^\top_{v\rightarrow va}z_{a\rightarrow va}/\tau)
\end{equation}
The $\mathrm{MILNCE}_{vt}(x)$ component is a variant of the $\mathrm{NCE}$ loss used to account modalities misalignment \cite{miech2020end}, and detailed in 
\begin{arxivPrint}\Cref{loss-function-appendix}.\end{arxivPrint}
\begin{confPrint}the supplemental materials.\end{confPrint}
$\mathcal{N}(x)$ is the set of negative modality pairs for the video $x$ in a given mini-batch, and $\tau$ is the temperature parameter. Utilizing MMV, we sought to confirm our hypothesis that an out-of-domain based network could be beneficial for better generalization.

\subsection{Deepfake detection downstream task}
\label{subsec:deepfake-detection-downstream}
To tackle the task of video deepfake detection, we propose \emph{FakeOut}, a fine-tuned model, pre-trained in the self-supervised manner detailed above, on data out-of-domain of the video deepfake detection downstream task.

\noindent
\textbf{Supervised downstream task.}
We used the multi-modal versatile backbone described in \Cref{subsec:backbone} to output modality representations. Given a video clip containing RGB frames and audio that was attained as a result of the enrichment process (\Cref{subsec:experiments-setup}), we obtain modality representations. The representations are concatenated to achieve a cross-modal representation of the clip. Then, we ingest these representation into a classification head. The latter outputs logits, followed by a log-softmax to achieve the final classification score. The score is used to determine whether the video is considered a deepfake (in the range $[0, 1]$, $0$ indicating \textit{real}). We focus on both the video and the audio modalities as their interaction proved to be beneficial for deepfake detection, rather than single modality approaches \cite{haliassos2022leveraging, zhou2021joint}. The full architecture of FakeOut is outlined in \Cref{fig:architecture}.

We use binary cross-entropy as the loss function for the fine-tuning phase: 
\begin{equation}
    \mathcal{L}(x)=y\mathrm{log}(\hat{y})+(1-y)\mathrm{log}(1-\hat{y})
    \label{eq:cross-entropy}
\end{equation}
Where $\hat{y}$ is the output of the log-softmax applied on the logits of $x$, and $y$ is the ground-truth label.
For every enriched video in the train-set, we randomly sample clips of 32 consecutive frames and their corresponding audio stream (more details in \Cref{subsec:experiments-setup}). The score yielded by our classifier for this clip is used in addition to the ground-truth label derived from the full video's label. Both backbone and classification head weights are not frozen during the fine-tuning process.

During inference time, we sample multiple, uniformly spaced clips from each pre-processed input video. The video-level score is the average of all clip scores.
If the face tracking pipeline (\Cref{subsec:data-preparation}) 
yielded multiple face tracks,
we use the max score out of all the tracks.

\subsection{Data preparation and pre-process}
\label{subsec:data-preparation}
As part of our data-centric approach, we found that data preparation is key to the success of our method. We created a pre-processing pipeline of data standardization, optional modality enrichment, face tracking, and augmentations. 

\noindent
\textbf{Standardization.} All videos pass through a standardization process aimed to mitigate dataset variability. Videos are re-encoded to 29 FPS with 44,100Hz audio sample rate.

\noindent
\textbf{Modality enrichment.} To enable multi-modal learning, we enrich the datasets with auditory information that was not included in their original version. The auditory enrichment process depends on specific dataset details, which are further explained in \Cref{subsec:experiments-setup}.

\noindent
\textbf{Face tracking.} We developed a face tracking pipeline which outputs sequences of tracked and aligned faces in a given video. We use full face crops, 
since models could benefit from these crops rather than the mouth region alone, 
specifically for multi-modal prediction tasks \cite{haliassos2022leveraging}. 
Our pipeline is robust to various video deepfake detection scenarios which appear in the commonly used datasets. 
For example, it can handle multi-person scenarios in which several people appear in the same video, while one or more might be forged. Specifically, the pipeline should distinguish between a multi-person scenario and a single person scenario with a crowded background. 

Given an input video, a coarse area of the face is obtained with MTCNN's bounding box detection. Bounding boxes are enlarged and then passed through MediaPipe face landmarks extraction process. We found that using MediaPipe alone for face detection on the original videos of most dataset at a high-quality compression rate does not deliver results as good as using our suggested process, more details are available in the supplemental materials. To ensure consistent faces within each frame sequence, we select the face bounding box with the highest IOU compared to the previous frame where a face was detected. Faces are aligned and centered using a similarity transformation of the detected face and of a reference mean face shape.

The pipeline outputs, for each video, one or multiple sequences (for multi-person scenarios). Each sequence consists of 224x224 RGB frames of a centered and aligned face. Corresponding audio is added if it either exists or obtained by the enrichment process (\Cref{subsec:experiments-setup}). See
\begin{arxivPrint}\cref{face-tracking-pipeline-appendix}\end{arxivPrint}
\begin{confPrint}supplemental materials\end{confPrint}
for further technical details. 

\noindent
\textbf{Augmentations.} During fine-tuning, we apply augmentations on each clip sampled from a video and ingested into the model. We follow the augmentation process in MMV pre-training, and therefore visual modality is augmented while audio modality is not. Horizontal flip and color augmentation are randomly applied as visual augmentations. No augmentations are applied during inference time. See 
\begin{arxivPrint}\cref{pre-process-appendix}\end{arxivPrint}
\begin{confPrint}supplemental materials\end{confPrint}
for more details.

\vspace{0.15cm}
As described above, our method includes a relatively standard paradigm (a self-supervised backbone followed by fine-tuning). Our contribution is \emph{not} this paradigm, which was applied to various other tasks. Instead, our contribution is in identifying that multi-modal out-of-domain data can be used for the backbone, and a technical contribution in our robust data processing pipeline, which lead to state-of-the-art results in cross-dataset deepfake detection. Next we report on experiment results and ablations.
\section{Experiments}
\label{sec:experiments}
In this section we evaluate the performance of our video deepfake detection method. First, we specify implementation details, then we describe the datasets used throughout the experiments. 
Several datasets are passed through our enrichment process which is also detailed in this section and used for evaluation. Then, we compare our method to state-of-the-art approaches, focusing on our ability to generalize well to unseen data and forgery techniques.
\subsection{Setup}
\label{subsec:experiments-setup}
\noindent
\textbf{Network architectures, hyper-parameters and optimization.}
For the backbone, we use Temporal Shift Module for the visual modality, based on ResNet50 (TSM-50) or ResNet50 with all channels doubled (TSM-50x2) \cite{lin2019tsm, kolesnikov2019revisiting}. We use ResNet50 for the audio modality. Backbones were pre-trained and made available by Alayrac~\etal~\cite{alayrac2020self}. Spatial and average pooling are applied at the last layer of the visual backbone to obtain the visual representation vector. 
Raw audio waveforms are transformed to log-MEL spectrograms with 80 bins. Spatial pooling is applied at the last layer of the backbone to obtain the audio representation vector. Visual modality's vectors are of dimension $d_v=$ 2048 for TSM-50 and $d_v=$ 4096 for TSM-50x2. Audio modality vectors are of dimension $d_a=$ 2048. After concatenation, cross-modality representation vectors of dimension 4096 and 6144 are created for TSM-50 and TSM-50x2 based backbones respectively. The classification head is a multi-layer perceptron implemented with 2 hidden layers. Adam optimizer is used with a cosine decay schedule with initial value of $1\mathrm{e}{-5}$ and $\alpha=$ 0.95.

\noindent
\textbf{Pre-training datasets.}
The backbone was pre-trained on out-of-domain multi-modal datasets which are invariant to the forgery detection task. The datasets used for pre-training are HowTo100M \cite{miech2019howto100m} and AudioSets \cite{gemmeke2017audio}. 

HowTo100M is a large-scale dataset of narrated videos with an emphasis on instructional videos with an explicit intention of explaining the visual content on screen. This dataset features 136 million video clips with captions sourced from YouTube videos. The AudioSet dataset is a large-scale collection of manually annotated audio events. It features 632 audio event classes and over 2 million human-labeled 10-second clips drawn from YouTube. 

Other self-supervised studies aiming to detect forged videos mainly rely on coupled pretext tasks that focus on the facial area and require data with facial attributes \cite{haliassos2022leveraging}, e.g., the LRW dataset \cite{Chung16}. In contrast, FakeOut's pre-training phase does not require data that features specific attributes such as clear human faces. Additionally, our reliance on out-of-domain data does not require any heavy pre-processing suck as landmark detection or ROI extraction during the pre-training phase. Also, collecting out-of-domain videos is easier opposed to domain-specific videos.

\noindent
\textbf{Forgery datasets.} We conducted experiments on several datasets, to test the ability of our system in versatile scenarios. 
We used FaceForensics++ (FF++) \cite{rossler2019faceforensics++} for training. FF++ contains 1000 real videos and 4000 fake videos generated by two face-swapping algorithms and two face reenactment algorithms. DeepFake~\cite{deepfake} and FaceSwap~\cite{faceswap} are used for face-swapping. NeuralTextures~\cite{thies2019deferred} and Face2Face~\cite{thies2016face2face} are used as reenactment algorithms. 
FF++ supplies three compression rates; we chose to use the high quality (HQ) version which is a middle-ground between raw and highly compressed video. %
Videos that are usually found in social media are in-line with this compression rate. 

For testing,
we use DeepFake Detection Challenge (DFDC) \cite{dolhansky2020deepfake}, featuring eight facial modification algorithms. DFDC test-set consists of 5000 videos. DeeperForensics (DFo) \cite{jiang2020deeperforensics} and FaceShifter (FSh) \cite{li2020advancing} are datasets created using the source videos of FF++ and a face-swapping technique. For evaluation on these 
datasets, we follow the FF++ train-test split.
DeepfakeTIMIT \cite{korshunov2018deepfakes} is an audio-visual dataset, in which fake videos are generated using a GAN-based approach, with no audio manipulations. The dataset comprises 640 videos, which were all used for testing in this work in their HQ version.
We also evaluated our method on FakeAVCeleb, another audio-visual deepfake dataset \cite{khalid2021fakeavceleb} that comprises both face-swapping \cite{faceswap, nirkin2019fsgan}, and facial reenactment \cite{prajwal2020lip} methods. To generate fake audio, a voice cloning tool \cite{jia2018transfer} was used. FakeAVCeleb features 500 real videos coming from Voxceleb2 \cite{chung2018voxceleb2} and 19,500 fake videos. We used 120 source videos and their forged versions as a test-set.
Additional common dataset in the field, is the Celeb-DF-v2 \cite{li2020celeb}, which is a video-only dataset. Relevant metadata to conduct the enrichment process described in \Cref{subsec:enrichment-process} is not provided, therefore it was not used for evaluation in this work.

\noindent
\textbf{Models for Comparison.} We compare our approach to other methods in the field; Xception \cite{rossler2019faceforensics++}, CNN-aug \cite{wang2020cnn}, and CNN-GRU \cite{sabir2019recurrent} use supervised learning both for the pre-training phase and the fine-tuning phase. Face X-ray \cite{li2020face} was trained on real images only, and reported relatively good generalization performance on unseen manipulation techniques. FTCN \cite{zheng2021exploring} consists of a fully temporal convolution network and a temporal transformer which seeks to leverage temporal coherence for video deepfake detection. AV DFD \cite{zhou2021joint} uses joint audio-visual representations in a supervised manner during train and test time. PCL + I2G \cite{zhao2021learning} learns to measure the patch-wise consistency of image source features boosted by an inconsistency image generator during training. LipForensics \cite{haliassos2021lips} is pre-trained on the LRW dataset in a supervised manner. 
RealForensics \cite{haliassos2022leveraging} is pre-trained in a self-supervised manner, also on the facial-related LRW dataset (with omitted labels). AUNet \cite{bai2023aunet} uses action-units relation transformer and tampered action-unit predictions. Results of other approaches on audio-visual datasets are taken verbatim from Haliassos \etal, Zhao \etal and Bai \etal
\cite{haliassos2022leveraging, zhao2021learning, bai2023aunet}.

\noindent
\textbf{Metrics.} 
We follow previous works \cite{haliassos2021lips, rossler2019faceforensics++, zheng2021exploring, zhou2021joint} and use area under the receiver operating characteristic curve (ROC-AUC). Metrics are calculated over video-level scores obtained during inference time as explained in \Cref{subsec:deepfake-detection-downstream}.

\begin{figure}
    \centering
    \includegraphics[width=\linewidth]{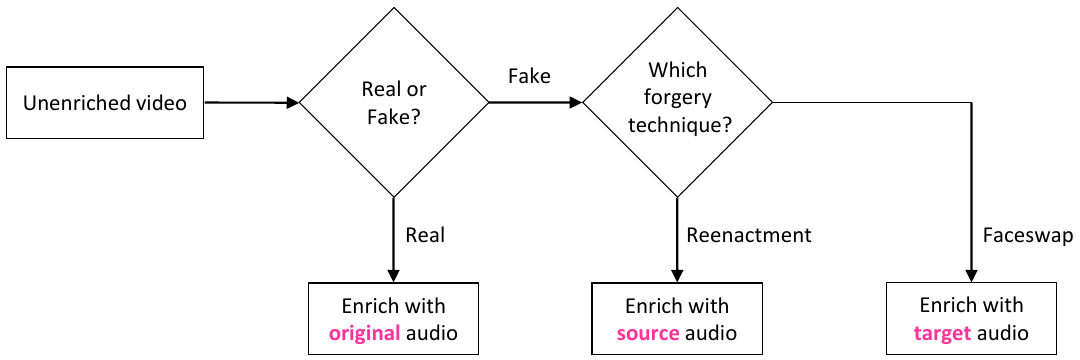}
    \caption{\textbf{Our enrichment process.} Each video in the FaceForensics++, DeeperForensics and FaceShifter datasets is enriched with the relevant audio file if it is available, according to this scheme.}
    \label{fig:enrichment}
    \vspace*{-5mm}
\end{figure}

\subsection{Enrichment process}
\label{subsec:enrichment-process}
We argue that the usage of all available modalities can yield benefit, thus we seek to enable utilization of the audio modality for fine-tuning. Therefore, 
we applied an enrichment process on the FF++, DFo and FSh datasets. Prior state-of-the art approaches in the field \cite{haliassos2021lips,haliassos2022leveraging,zheng2021exploring} used these datasets video modality only as they are provided this way in their official release. However, in addition to the silent videos, Rossler \etal\ \cite{faceforensics++github} reference the original YouTube videos that were used to create FF++, and specify the exact frame numbers taken. Since DFo and FSh are based on the same original videos, it enables our enrichment process.

During the enrichment process, each video is enriched with a corresponding audio stream extracted from the original videos available on YouTube. For forged videos, we distinguish between face-swapping techniques and reenactment techniques in order to enrich the video with the suitable audio stream, determined by the manipulation technique that was used for generation. Each manipulated video is a result of a \emph{source} sequence and a \emph{target} sequence; in face-swapping techniques, the source sequence is embedded into the target video while attempting to preserve the lip movement of the target video. In reenactment techniques, the source sequence is embedded into the target video while attempting to preserve the lip movement of the source video. 
Accordingly, we enrich each video with the relevant audio sequence that matches the lip movement as \Cref{fig:enrichment} explains.
We found incorrect frame mapping for several videos in the original dataset, and thus we manually verified the output of the enrichment process and used the un-enriched version for the videos with erroneous frame mappings. 
More details can be found in
\begin{arxivPrint}\cref{enrichment-process-validations}.\end{arxivPrint}
\begin{confPrint}the supplementals.\end{confPrint}

In \Cref{tab:datasets-enrichment} we show the available modalities in all datasets before and after the enrichment process. Fine-tuning our model with multi-modal data enables us to use all available modalities during inference time, which is crucial for generalizing well to multi-modal datasets such as DFDC, which may feature audio-visual mismatches in forged videos. The following cross-dataset experiments support this. 

\begin{table}[t]
  \centering
  \begin{adjustbox}{width=0.78\columnwidth}
  \tabcolsep=0.1cm
  \begin{tabular}{l c c c c}
  \toprule
  \multirow{2}[3]{*}{Dataset} & \multicolumn{2}{c}{Original} & \multicolumn{2}{c}{Enriched} \\
    \cmidrule(lr){2-3} \cmidrule(lr){4-5} & Visual     & Audio     & Visual    & Audio \\ \midrule
    Deepfake (FF++)                       & \faCheck   & \faTimes  & \faCheck  & \faCheck  \\ 
    NeuralTextures (FF++)                 & \faCheck   & \faTimes  & \faCheck  & \faCheck  \\ 
    Face2Face (FF++)                      & \faCheck   & \faTimes  & \faCheck  & \faCheck  \\ 
    FaceSwap (FF++)                       & \faCheck   & \faTimes  & \faCheck  & \faCheck  \\ 
    DFDC                                  & \faCheck   & \faCheck  & \faCheck  & \faCheck  \\
    DeeperForensics                       & \faCheck   & \faTimes  & \faCheck  & \faCheck  \\
    FaceShifter                           & \faCheck   & \faTimes  & \faCheck  & \faCheck  \\
    FakeAVCeleb                           & \faCheck   & \faCheck  & \faCheck  & \faCheck  \\
    DeepfakeTIMIT                           & \faCheck   & \faCheck  & \faCheck  & \faCheck  \\
    \bottomrule
  \end{tabular}
  \end{adjustbox}
  \caption{\textbf{Available modalities in forgery datasets.} Original column presents the available modalities natively in the commonly used face forgery dataset. Enriched column presents the available modalities after the enrichment process suggested in our work.
  }
  \label{tab:datasets-enrichment}
  \vspace*{-6mm}
\end{table}

\subsection{Evaluation}
\label{subsec:generalization-evaluation}
Due to the quick evolution in generation methods for forgery creation, we argue that the most important property of a deepfake detection system is generalization to unseen manipulation techniques and scenes. However, as Li \etal \cite{li2020face} observed in their work, classifiers experience a significant performance drop when applied to unseen manipulation methods. Therefore, we focus on evaluating cross-manipulation and cross-dataset tasks. Intra-dataset evaluation can be found in 
\begin{arxivPrint}\Cref{intra-dataset-evaluation}.\end{arxivPrint}
\begin{confPrint}the supplemental materials.\end{confPrint}
as we find it more saturated.
In all fine-tuning processes throughout the experiments, we maintain a balanced distribution of classes in the train-set by oversampling the minority class.

\begin{table}[t]
\centering
  \begin{adjustbox}{width=0.75\columnwidth}
  \begin{tabular}{>{\columncolor[gray]{0.95}}l c c c c}
    \toprule
    {Method} & {DF} & {FS} & {F2F} & {NT}\\ \midrule
    Xception \cite{rossler2019faceforensics++} & 93.9 & 51.2 & 86.8 & 79.7 \\
    CNN-aug \cite{wang2020cnn} & 87.5 & 56.3 & 80.1 & 67.8 \\
    Face X-Ray \cite{li2020face}& 99.5 & 93.2 & 94.5 & 92.5\\ 
    CNN-GRU \cite{sabir2019recurrent} & 97.6 & 47.6 & 85.8 & 86.6 \\
    LipForensics \cite{haliassos2021lips} & 99.7 & 90.1 & 99.7 & 99.1 \\
    AV DFD \cite{zhou2021joint} & 100.0 & 90.5 & 99.8 & 98.3 \\
    FTCN \cite{zheng2021exploring} & 99.9 & 99.9 & 99.7 & 99.2 \\
    PCL + I2G \cite{zhao2021learning} & 100.0 & 98.9 & 99.9 & 97.6 \\
    AUNet \cite{bai2023aunet} & 99.9 & 99.6 & 99.9 & 98.4 \\
    RealForensics \cite{haliassos2022leveraging} & 100.0 & 97.1 & 99.7 & 99.2 \\
    \midrule
    FakeOut & 99.5 & 97.6 & 98.9 & 91.9\\ 
    \bottomrule
  \end{tabular}
  \end{adjustbox}
  \vspace*{-2mm}
  \caption{\textbf{Cross-manipulation generalization.} Video-level ROC-AUC (\%) when testing on each manipulation method of FaceForensics++, fine-tuning on the remaining three methods. 
  Manipulation methods are Deepfakes (DF), FaceSwap (FS), Face2Face (F2F) and NeuralTextures (NT).
  }
  \label{tab:cross-manipulation-evaluation}
    \vspace*{-2mm}
\end{table}
\begin{table}[t]
    \setlength{\tabcolsep}{8pt}
  \centering
    \begin{adjustbox}{width=0.96\linewidth}
    \begin{tabular}{l c c c c c c}
    \toprule
    \multirow{2}[3]{*}{Method} & \multicolumn{3}{c}{Face-swapping} & \multicolumn{3}{c}{Reenactment} \\
    \cmidrule(lr){2-4} \cmidrule(lr){5-7}
    & DF & FS & Avg. & F2F & NT & Avg. \\
    \midrule
    FakeOut & 99.5 & 97.4 & 98.5 & 96.6 & 77.7 & 87.2\\
    \bottomrule
    \end{tabular}
    \end{adjustbox}
    \centering
    \vspace*{-2mm}
    \caption{\textbf{Cross-approach generalization.} Video-level ROC-AUC (\%) when fine-tuning on one deepfake generation approach (e.g. reenactment), while evaluating on an unseen one (e.g. face-swapping). Face-swapping methods are Deepfakes (DF) and FaceSwap (FS). Reenactment methods are Face2Face (F2F) and NeuralTextures (NT).}
    \label{tab:cross_manipulation_extreme}
    \vspace{-5mm}
\end{table}
\noindent
\textbf{Cross-manipulation generalization.}
To directly evaluate the performance of our system on unseen manipulations and forgeries, without taking into account new scenes, poses or light settings, we use the various manipulation types within the FF++ dataset. Concretely, we follow previous works and evaluate our model in a leave-one-out setting --- fine-tuning on three out of the four manipulation methods featured in the dataset, and testing on the remaining one. Results are shown in \Cref{tab:cross-manipulation-evaluation}. FakeOut achieves competitive results with other state-of-the-art methods in this task.

\noindent
\textbf{Cross-approach generalization.} This experiment aims to show the generalization ability of FakeOut when only one deepfake generation approach (e.g. reenactment) is seen during training. Then, we evaluate the model on an unseen deepfake generation approach (e.g. face-swapping). Results are shown in \Cref{tab:cross_manipulation_extreme}. This experiment is an expansion of the cross-manipulation task presented in \Cref{tab:cross-manipulation-evaluation} and was not reported by prior studies. It better measures a model's robustness to unseen deepfake generation techniques, and we encourage future studies to follow it.

\begin{table}[t]
  \centering
  \begin{adjustbox}{width=1.0\columnwidth}
  \begin{tabular}{>{\columncolor[gray]{0.95}}l >{\columncolor[gray]{0.95}}l c c c c c}
    \toprule
    {Method}                                      &        & {FSh}         & {DFo}          & {DFDC (fu.)}  & {DFDC (fi.)}   & {Avg.}   \\ \midrule
    Xception\cite{rossler2019faceforensics++}     &        & 72.0          & 84.5           & ---           & 70.9           & 75.8           \\
    CNN-aug \cite{wang2020cnn}                    &        & 65.7          & 74.4           & ---           & 72.1           & 70.7           \\
    Face X-Ray \cite{li2020face}                  &        & 92.8          & 86.8           & ---           & 65.5           & 81.7           \\
    CNN-GRU \cite{sabir2019recurrent}             &        & 80.8          & 74.1           & ---           & 68.9           & 74.6           \\
    Multi-task  \cite{nguyen2019multi}            &        & 65.5          & 77.7           & ---           & 68.1           & 70.4           \\
    DSP-FWA   \cite{li2018exposing}               &        & 65.5          & 50.2           & ---           & 67.3           & 61.0           \\
    LipForensics \cite{haliassos2021lips}         &        & 97.1          & 97.6           & ---           & 73.5           & 89.4           \\
    FTCN  \cite{zheng2021exploring}               &        & 98.8          & 98.8           & ---           & 74.0           & 90.5           \\
    PCL + I2G  \cite{zhao2021learning}               &        & ---          & 99.4           & 67.5           & ---           & ---           \\
    AUNet  \cite{bai2023aunet}               &        & ---          & ---           & 73.8           & ---           & ---           \\
    RealForensics \cite{haliassos2022leveraging}  &        & 99.7          & 99.3           & ---           & 75.9           & 91.6           \\
    \midrule
    FakeOut TSM-50                                & (V)    & 99.0          & 99.8           & 72.8          & 75.0           & 91.3           \\ 
    FakeOut TSM-50x2                              & (V)    & 99.5          & 99.9           & 73.9          & 75.8           & 91.7           \\ 
    FakeOut TSM-50                                & (V\&A) & 99.8          & 99.9           & 73.2          & 75.5           & 91.7           \\ 
    FakeOut TSM-50x2                              & (V\&A) & \textbf{99.8} & \textbf{99.9} & \textbf{75.1} & \textbf{76.4}  & \textbf{92.0}  \\ 
    \bottomrule
  \end{tabular}
  \end{adjustbox}
  \vspace*{-2.5mm}
  \caption{\textbf{Cross-dataset generalization.} Video-level ROC-AUC (\%) when fine-tuning on FaceForensics++ and testing on FaceShifter (FSh), DeeperForensics (DFo) and Deepfake Detection Challenge (DFDC) in its two versions, full and filtered test-set. V, A indicate the usage of video and audio modalities during FakeOut fine-tuning, respectively. Best results are in \textbf{bold}.
  }
  \label{tab:cross-dataset-evaluation}
  \vspace*{-5mm}
\end{table}
\noindent
\textbf{Cross-dataset generalization.}
This experiment most closely simulates in-the-wild scenarios, in which we cannot expect videos to come from the same distribution as the training set, both in terms of manipulation types, and other aspects such as lighting and photography style. To evaluate this, we fine-tune our model on the FF++ train-set, and measure performance on other audio-visual datasets: FaceShifter, DeeperForensics (enriched) and DFDC full test sets. We show the results of both training on FF++ original train-set, and the results of training on the enriched FF++ train set as described in \Cref{subsec:enrichment-process}. Also, in order to compare to other methods, we evaluate FakeOut on a filtered version of DFDC suggested by Haliassos~\etal~\cite{haliassos2021lips} as well. The subset consists of single-person videos without extreme light settings, poses or corruptions. Other audio-visual methods \cite{cheng2022voice, chugh2020not, zhou2021joint} that do not adhere to the cross-dataset generalization task settings were not included for comparison.
In \Cref{tab:cross-dataset-evaluation} we show that FakeOut achieves state-of-the-art performance, surpassing other methods.

When focusing on the most challenging dataset in the field, DFDC, FakeOut achieves state-of-the-art results on the filtered test-set.
Importantly, the performance does not drop significantly when evaluating on the full test-set which includes many challenging scenarios that were dropped in previous studies so far. We address that achievement to the robustness of our model and our face-tracking pipeline. Full DFDC test-set performance in this task was not reported by other methods, therefore left blank. We believe that our face-tracking pipeline will be useful for evaluation of future methods on the full DFDC test-set.

FakeAVCeleb and DeepfakeTIMIT are two additional audio-visual datasets, however, were not evaluated in these cross-dataset generalization settings by previous works \cite{haliassos2022leveraging, chugh2020not, cheng2022voice}, and do not provide an official test-set split. Therefore, we provide the results of FakeOut on these two datasets separately in \Cref{tab:additional-cross-dataset-evaluation}, and we encourage future studies to follow our suggested test-set splits, which 
\begin{arxivPrint} is published as a contribution as well.\end{arxivPrint}
\begin{confPrint}we will publish as another contribution upon publication.\end{confPrint}

\subsection{Representation analysis}
\label{subsec:representation_analysis}
\begin{figure}[t]
  \centering
  \includegraphics[width=\linewidth]  {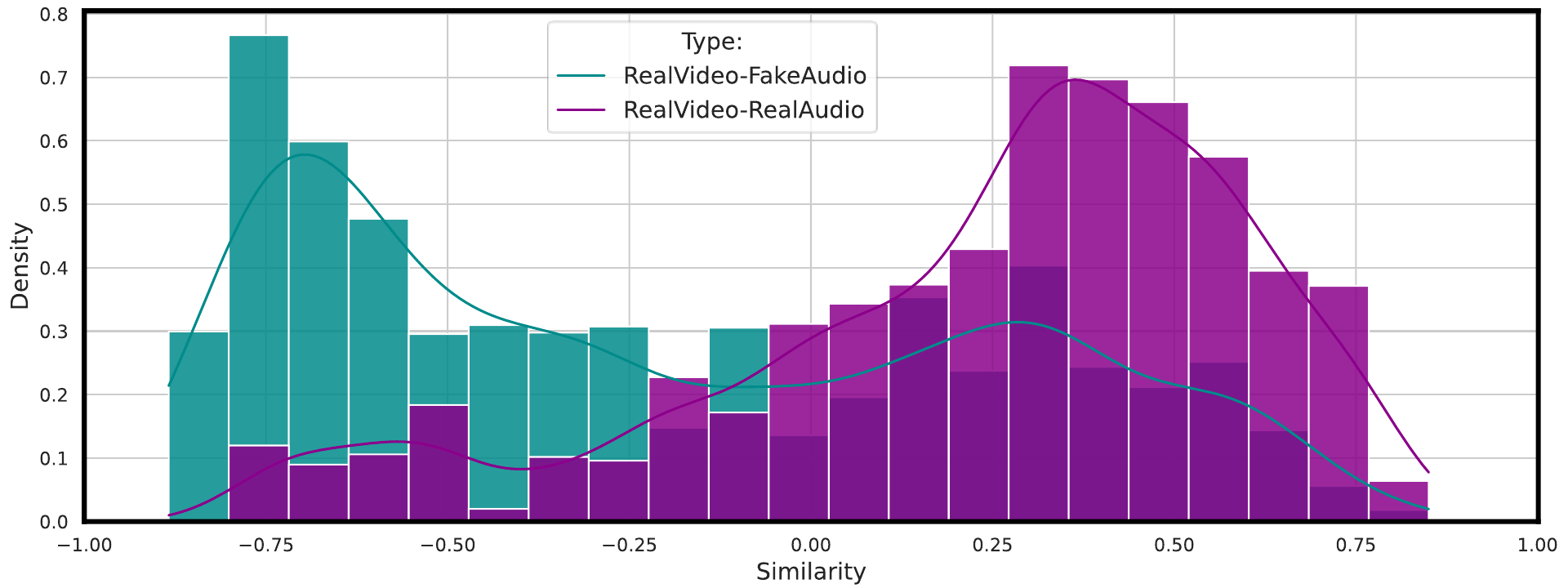}
  \vspace{-6mm}
  \caption{\textbf{Audio-visual features --- similarity.} Cosine similarity of the auditory and the visual feature-vectors extracted from the FakeAVCeleb's videos. The similarity values of the RealVideo-RealAudio type distribute around much higher values than in the RealVideo-FakeAudio case.}
  \label{fig:similarity-modalities}
  \vspace*{-5mm}
\end{figure}

To further investigate the quality of our system, we provide an analysis on the multi-modal representations extracted by FakeOut, as described in \Cref{fig:architecture}. In \Cref{fig:similarity-modalities} we analyze the embedded information in each modality of the videos in the FakeAVCeleb test-set. Although FakeAVCeleb is an unseen dataset during our training, FakeOut still extracts relevant information for the forgery detection task when audio modality is available. Additional analysis is provided in 
\begin{arxivPrint}\Cref{sec:additional_label_analysis}.\end{arxivPrint}
\begin{confPrint}the supplemental materials.\end{confPrint}

\begin{table}[t]
  \centering
  \begin{adjustbox}{width=0.87\columnwidth}
  \begin{tabular}{>{\columncolor[gray]{0.95}}l >{\columncolor[gray]{0.95}}l c c c}
    \toprule
    {Method}                                      &        & {FakeAVCeleb} & {DeepfakeTIMIT} \\ 
    \midrule
    FakeOut TSM-50                                & (V)    & 91.5          & 99.9            \\ 
    FakeOut TSM-50x2                              & (V)    & 92.3          & 99.9            \\ 
    FakeOut TSM-50                                & (V\&A) & 91.7          & 99.9            \\ 
    FakeOut TSM-50x2                              & (V\&A) & \textbf{93.4} & \textbf{99.9}   \\ 
    \bottomrule
  \end{tabular}
  \end{adjustbox}
  \vspace*{-2mm}
  \caption{\textbf{Cross-dataset generalization --- additional.} Video-level ROC-AUC (\%) when fine-tuning on FaceForensics++ and testing on FakeAVCeleb and DeepfakeTIMIT test-sets. Best results are in \textbf{bold}.}
  \label{tab:additional-cross-dataset-evaluation}
  \vspace*{-3mm}
\end{table}

\subsection{Ablation study}
\label{subsec:ablation-study}
We focus on two main components of FakeOut in the ablation study --- the adaption phase and the backbone.
Additional ablations can be found in \begin{arxivPrint}\Cref{sec:additional-ablation}.\end{arxivPrint}
\begin{confPrint}the supplemental materials.\end{confPrint}

\begin{figure}[t]
  \vspace*{-0.5mm}
  \centering
  \includegraphics[width=\linewidth]{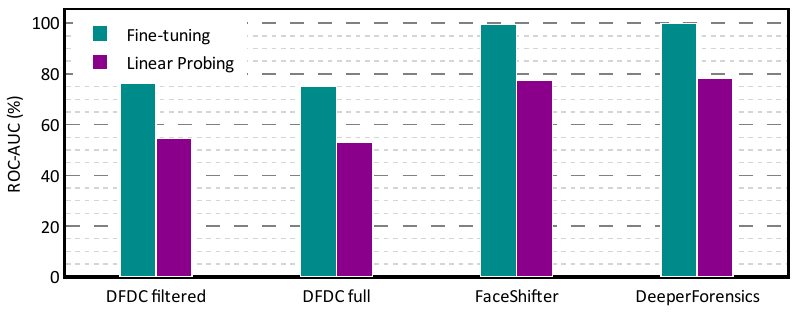}
  \vspace{-6mm}
  \caption{\textbf{Adaption ablation --- fine-tuning vs. linear probing.} We evaluate FakeOut on the cross-dataset generalization task, using two adaptation approaches to the video deepfake detection domain. FakeOut is trained on FF++ train-set. We compare fine-tuning the whole network vs. linear probing.}
  \label{fig:fine-tuning-vs-linear-probing}
  \vspace*{-2mm}
\end{figure}

\noindent\textbf{Adaption phase ablation.} Adapting the backbone to the deepfake detection task can be implemented in two main approaches --- fine-tuning and linear probing. Fine-tuning the full architecture of FakeOut requires more resources than linear probing, which only adapts the classification head. However, as we show in \Cref{fig:fine-tuning-vs-linear-probing}, fine-tuning the full FakeOut architecture is essential to obtain the 
generalization ability we reported in \Cref{sec:experiments} across all datasets. 
We compare the best performing version of FakeOut for each dataset.

\begin{figure}
  \vspace*{-2mm}
  \centering
  \includegraphics[width=\linewidth]{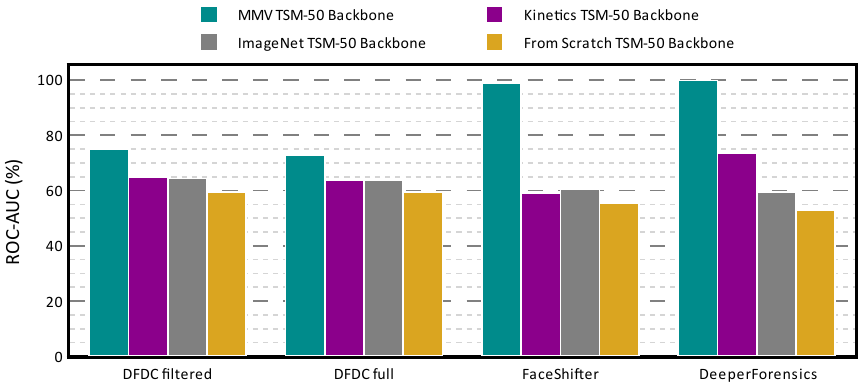}
  \vspace*{-6mm}
  \caption{\textbf{Backbone ablation.} To ablate the benefits of our self-supervised backbone trained in a multi-modal manner, we evaluate the cross-dataset task, using alternative TSM-50 backbones, trained in a supervised manner, or not pre-trained at all.}
  \label{fig:backbone-ablation}
\vspace*{-5mm}  
\end{figure}

\noindent\textbf{Backbone ablation.} In order to ablate the benefit of the backbone used as part of the FakeOut architecture, we incorporated different TSM-50 backbones, pre-trained either on the ImageNet \cite{deng2009imagenet} or the Kinetics \cite{kay2017kinetics} dataset, in a supervised manner on the visual modality, which is the only one available. Training from scratch with no pre-training is also compared. We used the same settings for fine-tuning, as we used to fine-tune the backbone of FakeOut. In \Cref{fig:backbone-ablation} we compare FakeOut with our proposed architecture, using video modality only, to FakeOut with the alternative backbones. We evaluate the models on the cross-dataset generalization task described in \Cref{subsec:generalization-evaluation}. FakeOut's performance surpasses its alternatives across all datasets.

\section{Limitations and future work}
\label{sec:limitations}
FakeOut struggles to classify well videos with \emph{extreme} post-processing, e.g. as DFDC features \begin{arxivPrint}(see \Cref{subsec:dog-mask-distractor}).\end{arxivPrint}\begin{confPrint}(see supplemental).\end{confPrint} These failure modes might indicate reduced robustness to post-processing which is becoming prevalent in social media. We believe new datasets can help alleviate this issue.

For future research, the audio modality could be leveraged even further to create more effective audio-visual representations. Applying audio augmentations might be beneficial, while also less explored in this field compared to the visual modality. Regarding architecture, incorporating attention-based mechanisms to the cross-modal representation creation may prove advantageous, as they achieve great performance in other video related tasks \cite{dosovitskiy2020image, shi2022avhubert}. 
\vspace{-2mm}

\section{Conclusion}
\label{sec:conclusions}
In this paper, we propose a new approach for video deepfake detection. We use both the video and audio modalities, and a self-supervised backbone that is invariant to our specific downstream task.
We show state-of-the-art results 
in deepfake detection generalization tasks on various datasets.
We attribute these results to our network architecture, a data-centric approach, a robust pipeline, and our enriched datasets, 
\begin{arxivPrint}which are all available on GitHub\footnote{\url{https://github.com/gilikn/FakeOut}} alongside the full training and inference code.\end{arxivPrint}
\begin{confPrint}which we will publish alongside the full training and inference code.\end{confPrint}
We hope FakeOut will push forward the field of deepfake detection, and help solve real-world problems in the era of fake news.
\begin{arxivPrint}

\textbf{Acknowledgements} This research was partially supported by the Israel Science Foundation (grant No. 1574/21).
\end{arxivPrint}

{\small
\bibliographystyle{ieee_fullname}
\bibliography{egbib}
}

\begin{arxivPrint}
\begin{appendices}
\appendix
\vspace{3mm}
In the next sections we provide additional details and results, further demonstrating our proposed method \textit{FakeOut}.
\section{Additional evaluations}
In addition to the generalization evaluation experiments presented throughout our work, in the following we share additional evaluations regarding the intra-dataset settings.

\subsection{Intra-dataset evaluation}
\label{intra-dataset-evaluation}
This experiment evaluates FakeOut on in-distribution data, i.e., both training and testing data are from the same distribution, featuring similar manipulation methods, and scenes. To simulate this, we fine-tune our model on the train-set of FaceForensics++ (FF++) \cite{rossler2019faceforensics++} and evaluate the model on the test-set of FaceForensics++. We can see in \Cref{tab:intra-dataset-evaluation} that FakeOut achieves competitive results with other state-of-the-art methods in terms of area under the ROC curve. We also see that FF++ is a saturated dataset, and the community will benefit from newer datasets for this evaluation. Regardless of our intra-dataset performance, we believe that inter-dataset performance is much more impactful for real-world detection scenarios, as explained in \Cref{subsec:generalization-evaluation}.
\begin{table}[ht]
  \centering
  \begin{tabular}{>{\columncolor[gray]{0.95}}l c c}
    \toprule
    {Method} & {Accuracy (\%)} & {ROC-AUC (\%)}\\ \midrule
    Xception\cite{rossler2019faceforensics++} & 97.0 & 99.3 \\
    CNN-aug \cite{wang2020cnn} & 96.9 & 99.1 \\
    Patch-based \cite{chai2020makes} & 92.6 & 97.2 \\
    Two-branch \cite{masi2020two} & --- & 99.1 \\
    Face X-Ray \cite{li2020face} & 78.4 & 97.8 \\
    CNN-GRU \cite{sabir2019recurrent} & 97.0 & 99.3 \\
    LipForensics \cite{haliassos2021lips} & 98.8 & 99.7 \\
    FTCN \cite{zheng2021exploring} & 99.1 & 99.8 \\
    RealForensics \cite{haliassos2022leveraging} & 99.1 & 99.8 \\
    \midrule
    FakeOut & 97.8 & 99.6 \\ 
    \bottomrule
  \end{tabular}
  \caption{\textbf{Intra-dataset evaluation.} We show the video-level performance, in terms of accuracy (\%) and ROC-AUC (\%), of several approaches, including FakeOut. All models are fine-tuned on the FaceForensics++ train-set and evaluated on the FaceForensics++ test-set.}
  \label{tab:intra-dataset-evaluation}
\end{table}

\begin{figure*}[t]
\centering
  \begin{subfigure}[b]{0.545\textwidth}
  \includegraphics[width=1.0\linewidth]{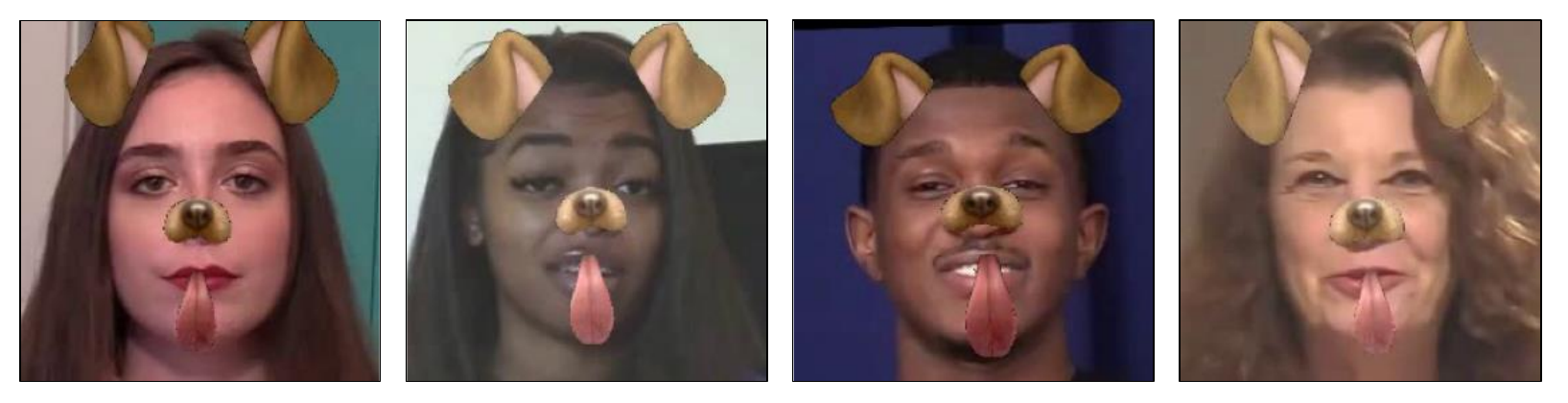}
  \caption{\textbf{Dog-mask distractor examples}}
  \label{fig:dog-mask-sub-figure}
  \vspace*{-2mm}
  \end{subfigure}
  \begin{subtable}[b]{0.45\textwidth}
  \centering
    \begin{adjustbox}{width=1.0\linewidth}
    \begin{tabular}{l l c c c c}
    \toprule
    \multirow{2}[3]{*}{Method} & & \multicolumn{2}{c}{DFDC (full)} & \multicolumn{2}{c}{DFDC (filtered)} \\
    \cmidrule(lr){3-4} \cmidrule(lr){5-6}
     & & w/ \faDog & w/o \faDog & w/ \faDog & w/o \faDog \\
    \midrule
    FakeOut TSM-50   & (V)    & 72.8 & \textbf{74.9} & 75.0 & \textbf{77.3} \\
    FakeOut TSM-50x2 & (V)    & 73.9 & \textbf{76.6} & 75.8 & \textbf{78.5} \\
    FakeOut TSM-50   & (V\&A) & 73.2 & \textbf{75.2} & 75.5 & \textbf{77.7} \\
    FakeOut TSM-50x2 & (V\&A) & 75.1 & \textbf{77.5} & 76.4 & \textbf{78.8} \\
    \bottomrule
    \end{tabular}
    \end{adjustbox}
    \centering
    \caption{\textbf{Generalization results, dropping dog-mask category}}
    \label{tab:dog_filter_figure}
    \vspace*{-2mm}
  \end{subtable}
\caption{\textbf{Dog-mask distractor category.} In \Cref{fig:dog-mask-sub-figure} we show frames sampled from videos of the dog mask distractor category in DFDC dataset. We consider this category of videos as an aggressive post-process, and argue it should be left out of the filtered DFDC test-set. In Tab. \ref{tab:dog_filter_figure} we report video-level ROC-AUC (\%) when fine-tuning on FaceForensics++ and testing on Deepfake Detection Challenge (DFDC) dataset in its two versions, filtered and full test-set, leaving out heavily post-processed examples of dog-mask filter.}
\label{fig:dog-mask-main-figure}
\vspace*{-3mm}
\end{figure*}

\begin{figure}
  \centering
  \includegraphics[width=0.45\textwidth]{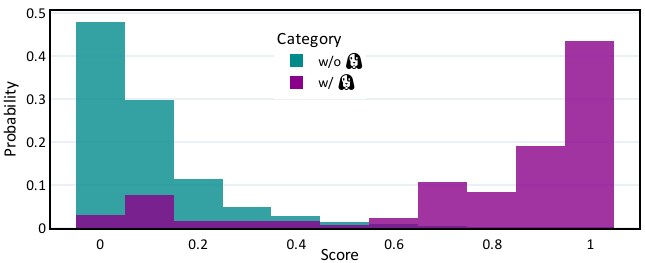}
  \caption{\textbf{Prediction scores distribution of \textit{real} videos in DFDC test-set.} This histogram shows the distribution of the prediction scores assigned by FakeOut to the \textit{real} videos in DFDC test-set. We distinguished between two categories; videos that the dog-mask distractor was applied on during post-processing and videos that it was not applied on. The prediction scores of the dog-mask distractor category are closer to $1$ although the videos are \textit{real}. However, regarding the rest of the test-set --- the prediction scores are closer to $0$ as we would expect from a well-performing model.
  }
  \label{fig:dog-mask-score-distribution}
\end{figure}
\subsection{Dog-mask distractor}
\label{subsec:dog-mask-distractor}
Investigating our failure modes, we found 215 videos in the DFDC test-set, for which a dog-mask filter was applied (\Cref{fig:dog-mask-sub-figure}). Crucially, these videos are marked as \emph{real} in the dataset. We argue that these videos should be evaluated separately, since they can be considered fake, depending on context. In \Cref{tab:dog_filter_figure} we show that our state-of-the-art results on DFDC improve further once these videos are removed from the test-set. The dog-mask distractor category in DFDC test-set was challenging for our model. If this category is left out of the test-set, the ROC-AUC scores are significantly improved as our experiments have shown. As part of our error analysis, we explored the prediction scores assigned to all videos labeled as \textit{real} in the DFDC test-set. The distributions of the prediction scores are presented in \Cref{fig:dog-mask-score-distribution}. These histograms clearly show the significant difference between the distribution of the dog-mask distractor category compared to the rest of the test-set. This exploration supports our argument that this category behaves differently from the rest of the test-set, as many video of it are marked as \textit{real} in the DFDC dataset, even though they are aggressively post-processed. This comparison motivated us to evaluate the dog-mask distractor category separately.
\label{dog-mask-distractor}

\section{Fine-tuning implementation details}
\subsection{Pre-process}
\label{pre-process-appendix}
As part of the pre-processing pipeline during the fine-tuning phase, we apply augmentations on each clip that was randomly sampled from the videos used for fine-tuning. For color augmentation we randomize hue (max delta=$1/5$) and brightness (max delta=$32/255$). We randomly flip clips horizontally with uniform distribution. Lastly, we ensure that the RGB values of each frame are in $[0, 1]$.

\subsection{Optimization}
During the adaption phase of FakeOut to the video deepfake detection domain, we fine-tune our network for $6$ epochs, chosen by the validation set, using Adam optimizer with parameters $\beta_1=0.9$, $\beta_2=0.999$ and $\epsilon=1\mathrm{e}{-8}$.

\section{Text modality}
\subsection{Pre-processing \& backbone}
The text modality is used solely during the pre-training phase via MMV, by Alayrac \etal \cite{alayrac2020self}. Automatic Speech Recognition was used on the videos to extract discrete word tokens and form the text-modality. These tokens pass through a standardization process detailed in the work of Alayrac \etal. Then, the processed tokens pass through the backbone of the text-modality. This backbone consists of several stages --- Word2Vec \cite{mikolov2013efficient} embeddings of the tokens are obtained, then the embeddings pass through a linear layer with a non-linear activation function. Lastly, max-pooling layer is applied on the learned representations. \gili{Added the following:} We used the pre-trained MMV backbone pre-trained on video, audio and text modalities due to the best performance reported by Alayrac \etal on various visual recognition tasks, achieved by using the all three modalities. No other backbone, omitting text-modality during pre-training, was made accessible by Alayrac \etal, therefore no ablation study could be performed regarding necessity of the text modality.

\begin{figure*}[t]
    \centering
    \includegraphics[width=\linewidth]{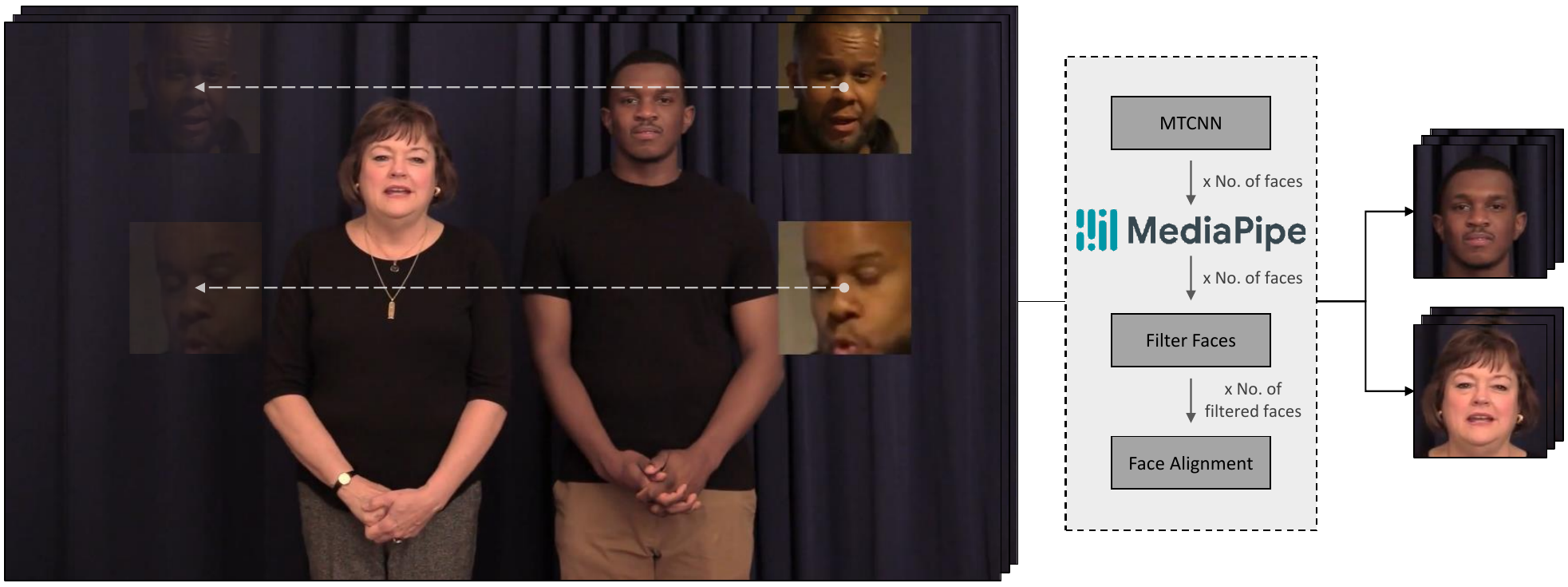}
    \caption{\textbf{Face tracking pipeline overview.} Every video passes through our face tracking pipeline. The pipeline consists of a few main stages. First, face bounding boxes are detected using MTCNN. Then, landmarks of each face in the bounding boxes are detected using MediaPipe. We filter out the irrelevant faces based on our logic to handle crowded background scenarios and distractors.
    Then, each relevant face is aligned using the detected landmarks and the reference mean face presented in \Cref{fig:reference_mean_face}. In the example shown above, from the DFDC dataset, two moving image distractors were added to the video during post-processing. The trajectories indicate the paths the distractors pass throughout the video (dashed arrows). Despite the challenging multi-person synthetic distractors, our face tracking pipeline successfully outputs two sequences of the aligned \emph{relevant} faces.}
    \label{fig:face_tracking_pipeline}
\end{figure*}
\subsection{Loss function}
\label{loss-function-appendix}
A contrastive loss function is used during the pre-training phase, via MMV. One component incorporated in the contrastive loss function, when the text modality is available, is the $\mathrm{MILNCE}$ loss. The $\mathrm{MILNCE}$ loss accounts for cross modality misalignment. Usually, the correspondence between the text extracted from a speaking narrator in a video and what is actually happening in the video is much weaker compared to the correspondence of the video and audio modalities. Therefore, Alayrac \etal chose to tackle the misalignment between the video and the text using this loss function component.
This component replaces the term $\exp(z^\top_{v\rightarrow vat}z^{ }_{t\rightarrow vat}/\tau)$ in the standard $\mathrm{NCE}$ loss equation by a sum of scores over positive text candidates: $\sum_{z\in\mathcal{P}(x)}\exp(z^\top_{v\rightarrow vat}z^{ }_{t\rightarrow vat}/\tau)$. This way, it considers multiple video-text candidates as positive samples. $\mathcal{P}(x)$ is a set formed from temporally close narrations yielding potential positives. The full formulations are as follows:
\begin{equation}
\label{eq:milnce}
\begin{split}
    &\mathrm{MILNCE}_{vt}(x) = \\ &-log \left( \frac{\sum_{x\in\mathcal{P}(x)}\mathrm{VT}(f(x))}{\sum_{x\in\mathcal{P}(x)}\mathrm{VT}(f(x)) + \sum_{x'\sim\mathcal{N}(x)}\mathrm{VT}(f(x'))}\right)
\end{split}
\end{equation}
\begin{equation}
    \label{eq:vat}
    \mathrm{VT}(z) = \exp(z^\top_{v\rightarrow vat}z^{ }_{t\rightarrow vat}/\tau)
\end{equation}

\section{Face tracking pipeline}
\label{face-tracking-pipeline-appendix}
Given an input video, during fine-tuning and inference time, the processes it passes through as part of the face tracking pipeline are as follows:
\begin{enumerate}
   \item \textbf{Coarse location of faces.} We find bounding boxes of detected faces with high confidence in each frame using the MTCNN algorithm. Bounding boxes are enlarged, i.e., height and width are multiplied by 1.8. When multiple faces are detected in a video, we use the multiple-faces mode. In this mode, we find the areas in the video that each face is located in at the beginning of the video. We assume that in this scenario, the coarse location of a single face would not change significantly. The last assumption does not hold for the single-face scenario, thus not used in that mode. To avoid detecting face-like distractors that intrude the area of a relevant face, we make sure that the detected face in the current frame is close in size to the last face detected in previous frames. If face bounding boxes are not found in a given frame, we interpolate the bounding boxes based on the ones from temporally close frames in which faces were indeed detected.
   \item \textbf{Landmark extraction.} We crop each frame according to the detected bounding boxes from the previous stage, and we process the cropped frames with the FaceMesh object via MediaPipe to extract landmarks. The combination of MTCNN in the previous stage and FaceMesh by MediaPipe handles better challenging scenarios, such as presented in \Cref{fig:mtcnn_mediapipe_combination} The FaceMesh object might extract landmarks of more than one face in a given cropped frame, if several faces exist. This is common in videos featuring crowded backgrounds.
   \item \textbf{Face filter.} We validate whether the landmarks of the detected face in a given frame are of the same face detected in previous frames choosing the face with max intersection-over-union (IOU) of the two landmark areas. Hence, we keep the landmarks of the main face throughout the frame sequence and filter the others, assuming it is the relevant one that was also detected by the MTCNN algorithm.
   \begin{figure}[t]
    \centering
    \includegraphics[width=\linewidth]{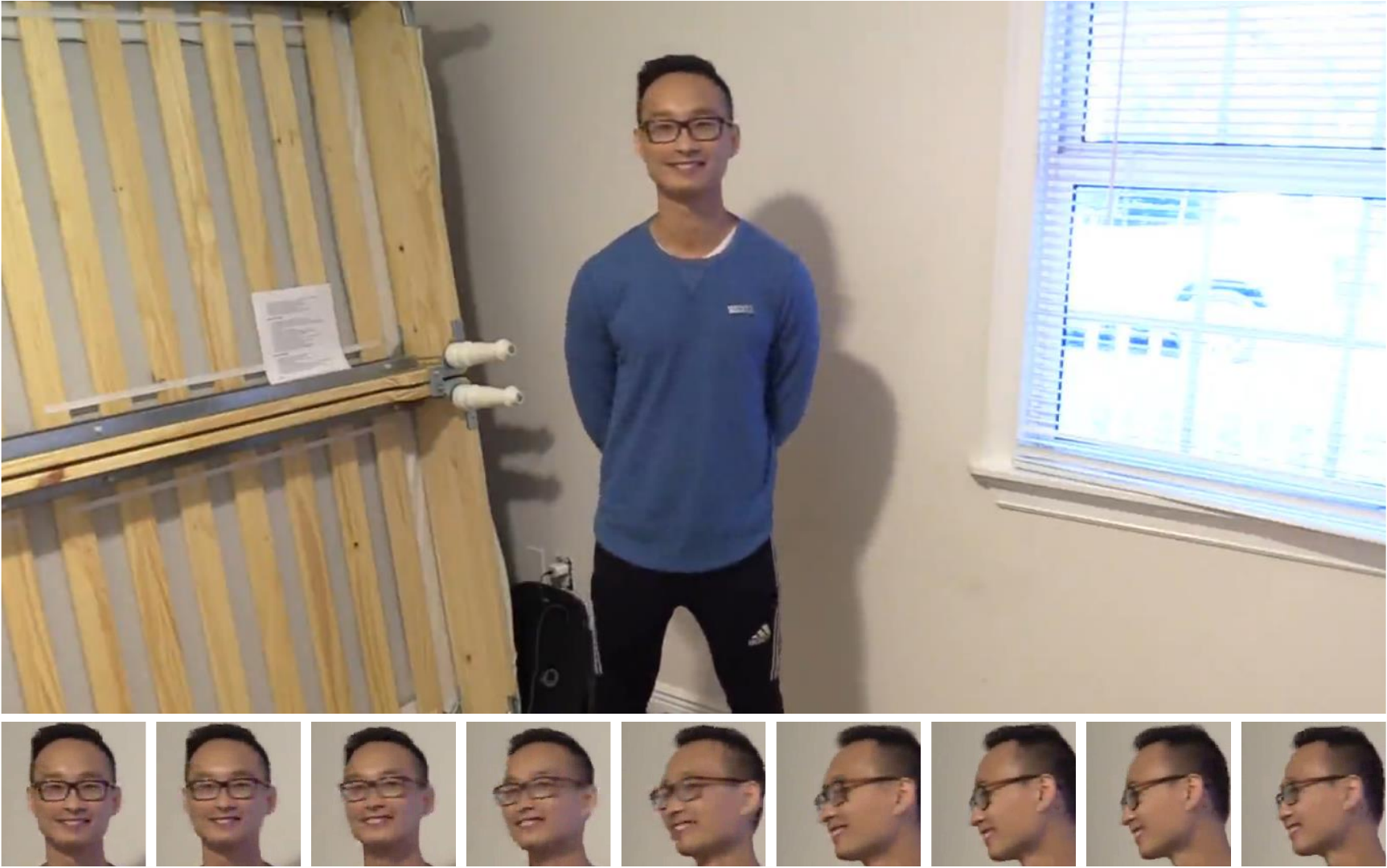}
    \caption{\textbf{Challenging face detection scenario.} In the presented example, we showcase a challenging scenario, in which a person stands in a distance from the camera, in a low quality video. The person turns his head, looking to the sides throughout the video. The movement, the distance and the video quality challenges MediaPipe to handle this scenario well, and it \emph{fails to detect} the face. Incorporating the pre-stage of detecting bounding boxes by MTCNN, enables us to track the face and cope with this scenario.}
    \label{fig:mtcnn_mediapipe_combination}
    \vspace{-4mm}
\end{figure}
   \item \textbf{Face alignment.} Landmarks are used for face alignment. If no landmarks were detected in the previous stage, we interpolate them from temporally adjacent frames. We apply a sliding mean filter of size $5$ to smooth the landmark values across time. Then, we compute the similarity transformation matrix between the $5$ key landmarks detected in the frame and the corresponding landmarks in the reference mean face as presented in \Cref{fig:reference_mean_face}. The transformation matrix is applied on the cropped frame to obtain an aligned face in the center of the frame of size $224$x$224$. 
\end{enumerate}
A schematic overview of the pipeline is presented in \Cref{fig:face_tracking_pipeline}.

\section{Additional model analysis}
\label{sec:additional_label_analysis}
In addition to the audio-visual analysis presented in \Cref{subsec:representation_analysis}, we elaborate in the following subsections with more visualizations and analysis scenarios.
\noindent
\begin{figure}[t]
  \centering
  \includegraphics[width=0.48\textwidth]{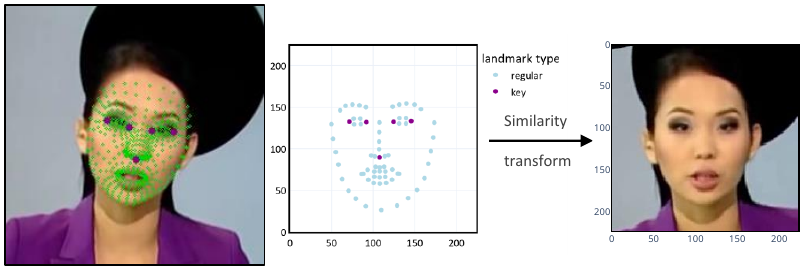}
  \caption{\textbf{Similarity transform with reference mean face.} Landmarks of the reference mean face used for alignment. For every frame, we calculate the similarity transformation of the detected key landmarks of the face and the key landmarks of the reference mean face (dark-magenta colored).
  }
  \label{fig:reference_mean_face}
  \vspace{-3mm}
\end{figure}
\subsection{Visualization of multi-modal representations}
In \Cref{fig:umap-visualization} we show an analysis of the representation vectors FakeOut produces for video clips from the FaceShifter datasets. The $2$D visualization presents the feature vectors after a UMAP \cite{mcinnes2018umap} dimensionality reduction. There is a clear differentiation between the \emph{fake} and \emph{real} videos, in addition to the ability of FakeOut to produce clusterable feature vectors of the sampled clips of each video.

\begin{figure}[t]
\centering
  \includegraphics[width=1\linewidth]{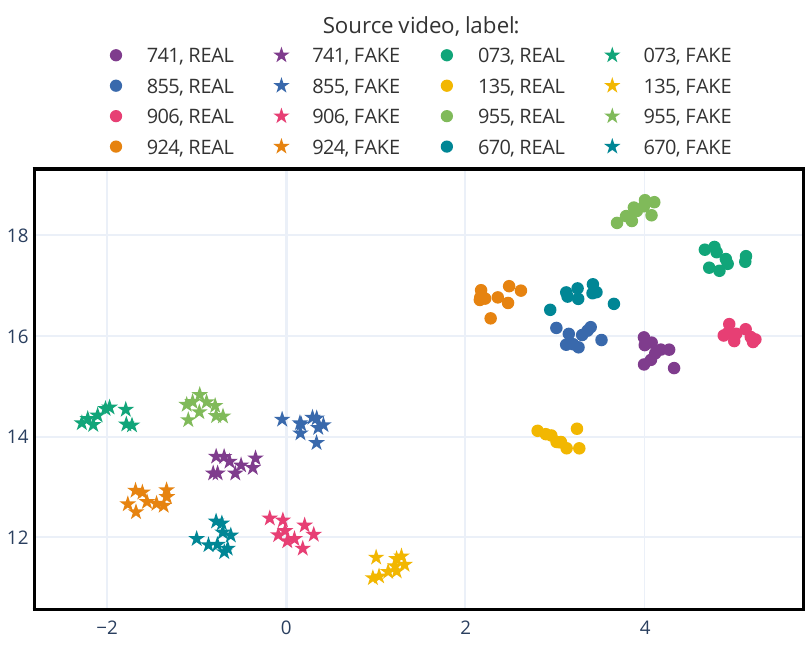}
  \vspace{-7mm}
  \caption{\textbf{UMAP dimensionality reduction of FakeOut's vectors.}
  We visualize the audio-visual feature vectors of FakeOut, on the FaceShifter use-case after enrichment, using UMAP \cite{mcinnes2018umap} dimensionality reduction. This visualization shows the ability of our system to separate the \emph{fake} videos from the \emph{real} ones. The plot visually supports the high performance FakeOut achieves on the FaceShifter cross-dataset evaluation.}
  \label{fig:umap-visualization}
  \vspace{-1mm}
\end{figure}
\begin{figure}[t]
\centering
  \begin{subfigure}[b]{0.5\linewidth}
  \includegraphics[width=\linewidth]{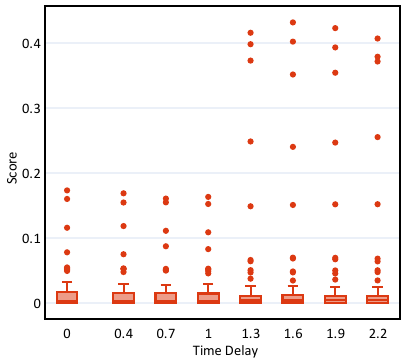}
  \caption{\textbf{Effect of time-delay.}}
  \label{fig:audio_shift_box}
  \end{subfigure}
  \begin{subfigure}[b]{0.49\linewidth}
  \includegraphics[width=\linewidth]{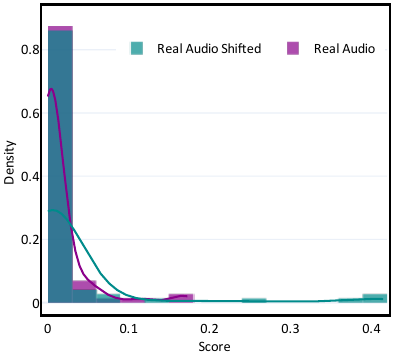}
  \caption{\textbf{Scores distribution.}}
  \label{fig:audio_shift_histogram}
  \end{subfigure}
\caption{\textbf{Audio shift analysis.} In this analysis, we investigate the ability of our model in handling a temporal shift of \emph{real} videos with \emph{real} audio. We used the videos from the RealVideo-RealAudio category of FakeAVCeleb's test-set as baseline. Then, we shifted the audio of each video by different time delays. In \Cref{fig:audio_shift_box} we present the box-plots of the model's scores for different time delays applied to each video. In \Cref{fig:audio_shift_histogram}, we can observe the score-distribution change after shifting the audio by 1.3s.}
\label{fig:audio_shift_analysis}
\vspace*{-4mm}
\end{figure}

\subsection{Audio shift analysis}
In this experiment, we examine the ability of FakeOut to handle audio shifts in \emph{real} videos. We used the RealVideo-RealAudio category provided in the FakeAVCeleb test-set~\cite{khalid2021fakeavceleb}. 
We applied a temporal shift to each video's audio, so that the audio does not correctly correspond to the video frames. 
Several values of time-delay were explored as can be observed in \Cref{fig:audio_shift_box}. 
After shifting, FakeOut outputs a score for each video. 
As we show in the figure, FakeOut outputs a higher score as time-delay increases. 
  
However, this behavior does not increase monotonically over the delay values, and does not apply to all videos. 
The videos which are significantly affected, are affected at a time-delay of 1.3s or higher. 
This behavior is in line with the characteristics of real videos --- many real videos have perfectly aligned audio, some have small alignment issues (e.g., due to diverging audio and video processing pipelines), but hardly any real videos have large misalignments in the wild. 
Whether we consider misalignments as \emph{real} or \emph{fake} is application dependent, and such misalignments can be added to the training set if we want to consider them \emph{fake}.

\section{Additional ablations}
\label{sec:additional-ablation}
To further ablate our system, in addition to the ablation study presented in \Cref{subsec:ablation-study}, in this section we analyze the inference process, analyze the confidence of the model, and compare our backbone to an alternative self-supervised backbone.
\noindent
\subsection{Number of clips sampled in test time} Throughout the evaluations, FakeOut assigns a prediction score for each clip. To obtain video-level predictions, we sample several clips from each video in test time and average the scores. Clips are non-overlapping (if video length permits). In \Cref{fig:number_of_windows_supp} we report ROC-AUC scores when training on FF++ and testing on DFDC, FaceShifter and DeeperForensics for both TSM-50 and TSM-50x2 as backbones.
Performance increases, across all datasets, as the number of clips sampled from each video in test time increases. We hypothesize that FakeOut will perform even better if more clips were sampled, but due to GPU limitations we use a maximum of 9 sampled clips per video.

\begin{figure}[ht]
  \centering
  \includegraphics[width=0.47\textwidth]{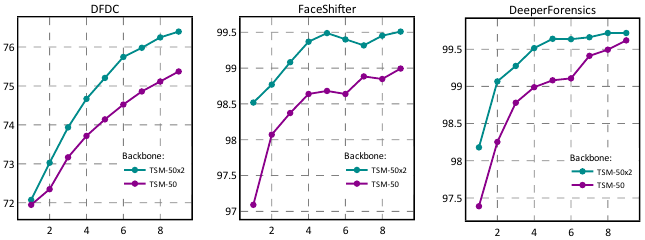}
  \caption{\textbf{Effect of number of clips sampled during test time across datasets.} ROC-AUC (\%) as a function of number of clips sampled from each video at test time. 
  We show that ROC-AUC increases with the number of sampled clips across the DFDC, FaceShifter and DeeperForensics datasets.
  }
  \label{fig:number_of_windows_supp}
  \vspace*{-3mm}
\end{figure}
\subsection{Fine-tuning vs. linear probing}
Here we show how fine-tuning as an adaption approach performs significantly better compared to linear probing, in terms of confidence level. In \Cref{fig:confidence_supp} we see the difference in prediction score distributions between the two approaches, in which the fine-tuned model is much more certain about its predictions compared to the other. This insight also encourged us to use fine-tuning as the adaption approach for FakeOut.
\begin{figure}
  \centering
  \includegraphics[width=0.47\textwidth]{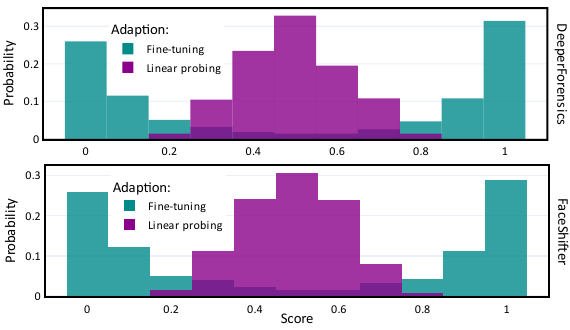}
  \caption{\textbf{Model confidence --- fine-tuning vs. linear probing.} The histograms show the distribution of scores our model outputted, after adapting it to the video deepfake domain. 
  The distribution of scores after fine-tuning and linear probing are colored in dark-cyan and dark-magenta respectively. We observe that the scores after fine-tuning tends to be closer to $0$ or $1$, while the scores after linear probing are around $0.5$. Therefore, 
  the model is 
  more confident about its predictions after fine-tuning.
  We use fine-tuning as the adaptation approach in FakeOut.
  }
  \label{fig:confidence_supp}
  \vspace*{-4mm}
\end{figure}
\begin{figure}[t]
\centering
  \vspace*{-2mm}
  \includegraphics[width=1\linewidth]{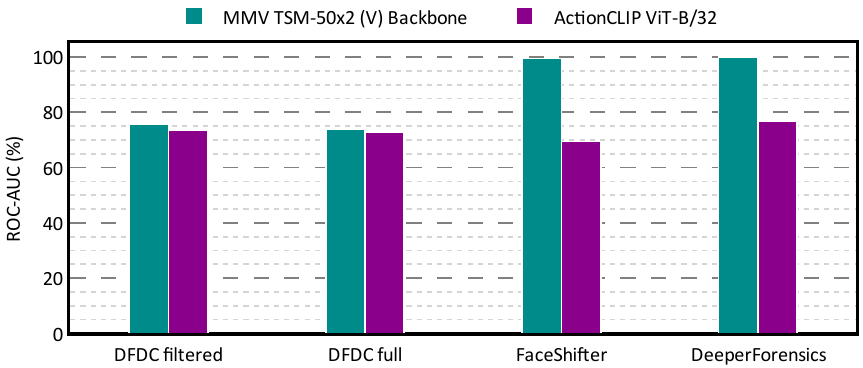}
  \caption{\textbf{MMV vs. CLIP.}
  We examine the generalization ability of FakeOut, which is based on the MMV backbone, in comparison to another robust self-supervised model, CLIP. Specifically, we fine-tuned a modified version of CLIP suited for video-recognition --- ActionCLIP. Equivalently to FakeOut (TSM-50x2 V), ActionCLIP was fine-tuned on FaceForensics++ train set, using frames-only, and was evaluated in the cross-dataset settings on the same datasets presented throughout the paper.}
  \label{fig:mmv_vs_clip}
  \vspace*{-2mm}
\end{figure}

\subsection{Self-supervised method --- MMV vs. CLIP}

We measured the performance of FakeOut versus a fine-tuned ActionCLIP model \cite{wang2021actionclip}, which is a modification of the self-supervised model CLIP \cite{radford2021learning}, suited for video-recognition tasks.
We fine-tuned a version of ActionCLIP based on ViT-B/$32$ architecture, which was pre-trained on publicly available image-caption data in a self-supervised manner. 
The pre-training was done through a combination of crawling a handful of websites and using commonly-used pre-existing image datasets such as YFCC100M \cite{thomee2016yfcc100m} which contains a total of $100$ million media objects. Results are shown in \Cref{fig:mmv_vs_clip}. 
We can see that even when FakeOut does not incorporate the enriched audio modality, it still overtakes a CLIP Vision Transformer trained in a self-supervised approach and fine-tuned on the same processed data we obtained by our pipeline. 
Modifying ActionCLIP to the video deepfake detection task was enabled by their open-source code, and our fine-tuned checkpoint will also be published for reproducibility.

\section{Enrichment process validations}
\label{enrichment-process-validations}
As part of our work, we enrich the silent videos of FaceForensics++ with the audio sequences from the original videos used to create the dataset. The exact frame mappings and YouTube URLs were made available by Rossler \etal\ \cite{faceforensics++github}. After a manual validation process, we found that only 737 videos, out of the 1000 source videos that the dataset features, had an available corresponding YouTube video on the provided URL. After the enrichment process we manually verified that the audio addition aligns perfectly with the frame sequences. We found that 36 videos do not have a correct frame mapping. Hence, only 701 of the videos that were verified to have matching audio were used enriched, while the others were used un-enriched.

\end{appendices}
\end{arxivPrint}

\end{document}